\documentclass[letterpaper, 10 pt, conference]{ieeeconf}  

\IEEEoverridecommandlockouts                              

\usepackage[utf8]{inputenc}
\usepackage[T1]{fontenc}
\usepackage{amsmath,amssymb,amsfonts}
\usepackage{graphicx}
\usepackage{textcomp}
\usepackage{xcolor}
\DeclareUnicodeCharacter{0308}{\"{}}
\usepackage{cite}
\usepackage{xurl}
\usepackage{subcaption}
\usepackage[table]{xcolor}   
\usepackage{colortbl}
\usepackage{booktabs}
\usepackage{multirow}
\usepackage{mathrsfs}
\usepackage[ruled,noend,linesnumbered]{algorithm2e}

\usepackage[colorlinks,linkcolor=black,anchorcolor=black,urlcolor=black,citecolor=black]{hyperref}

\usepackage[normalem]{ulem}
\useunder{\uline}{\ul}{}

\title{
\LARGE \bf TopAY: Efficient Trajectory Planning for Differential Drive\\Mobile Manipulators via Topological Paths Search\\and Arc Length-Yaw Parameterization
}

\begin{document}

\author{Long Xu$^{\dagger,1,2}$, Choilam Wong$^{\dagger,2}$, Mengke Zhang$^{1,2}$, Junxiao Lin$^{1,2}$, Jialiang Hou$^{1,2}$ and Fei Gao$^{1,2}$
\thanks{$^{\dag}$Indicates equal contribution.}
\thanks{$^{1}$State Key Laboratory of Industrial Control Technology, Zhejiang University, Hangzhou 310027, China. \textit{Corresponding author: Jialiang Hou}}
\thanks{$^{2}$Huzhou Institute of Zhejiang University, Huzhou 313000, China.}
\thanks{E-mail: {\tt\small \{gaolon, fgaoaa\}@zju.edu.cn}}
}

\maketitle
\thispagestyle{empty}
\pagestyle{empty}

\begin{abstract}
Differential drive mobile manipulators combine the mobility of wheeled bases with the manipulation capability of multi-joint arms, enabling versatile applications but posing considerable challenges for trajectory planning due to their high-dimensional state space and nonholonomic constraints. This paper introduces TopAY, an optimization-based planning framework designed for efficient and safe trajectory generation for differential drive mobile manipulators. The framework employs a hierarchical initial value acquisition strategy, including topological paths search for the base and parallel sampling for the manipulator. A polynomial trajectory representation with arc length–yaw parameterization is also proposed to reduce optimization complexity while preserving dynamic feasibility. Extensive simulation and real-world experiments validate that TopAY achieves higher planning efficiency and success rates than state-of-the-art method in dense and complex scenarios. The source code is \textbf{released} at \href{https://github.com/TopAY-Planner/TopAY}{\textcolor{magenta}{https://github.com/TopAY-Planner/TopAY}}.
\end{abstract}

\section{Introduction}
\label{sec:Introduction}
Differential drive mobile manipulator (DDMoMa), comprising multi-joint manipulator(s) mounted on a differential drive base (DDB), integrates rich manipulation ability of manipulators and mobility of wheeled robots. This synergistic combination enables versatile application in diverse domains, including industrial manufacturing, healthcare, agriculture, etc\cite{survey}. For DDMoMa to perform complex tasks, autonomous navigation and obstacle avoidance are fundamental.


Prevailing methods \cite{zzd_icra,rampage,trajopt} rely on numerical optimization to obtain safe, dynamically feasible, and optimal trajectories under certain criteria. The initial values for optimization are commonly acquired by a combination of sampling- and search-based methods\cite{sampling, hybridA}. Despite demonstrating capability of real-time planning, their efficiency degrade significantly in complex scenarios, due to the difficulty of handling nonholonomic constraints of the DDB and high-dimensional state space of DDMoMa.


Due to the exponential growth of space volume with dimensionality, the combined state space of a DDB and manipulator(s) is significantly larger than their individual state spaces. This elevates the complexity of initial value acquisition and increases number of optimization variables for trajectory optimization. Critically, the presence of manipulator(s) necessitates 3D collision checking, rendering the simplified 2D planning methods common to indoor wheeled robots inadequate. Compounding this, the mobility of the base demands collision checking to be performed over a larger physical space.

\begin{figure}[t]
    \centering
    \includegraphics[width=1.0\columnwidth]{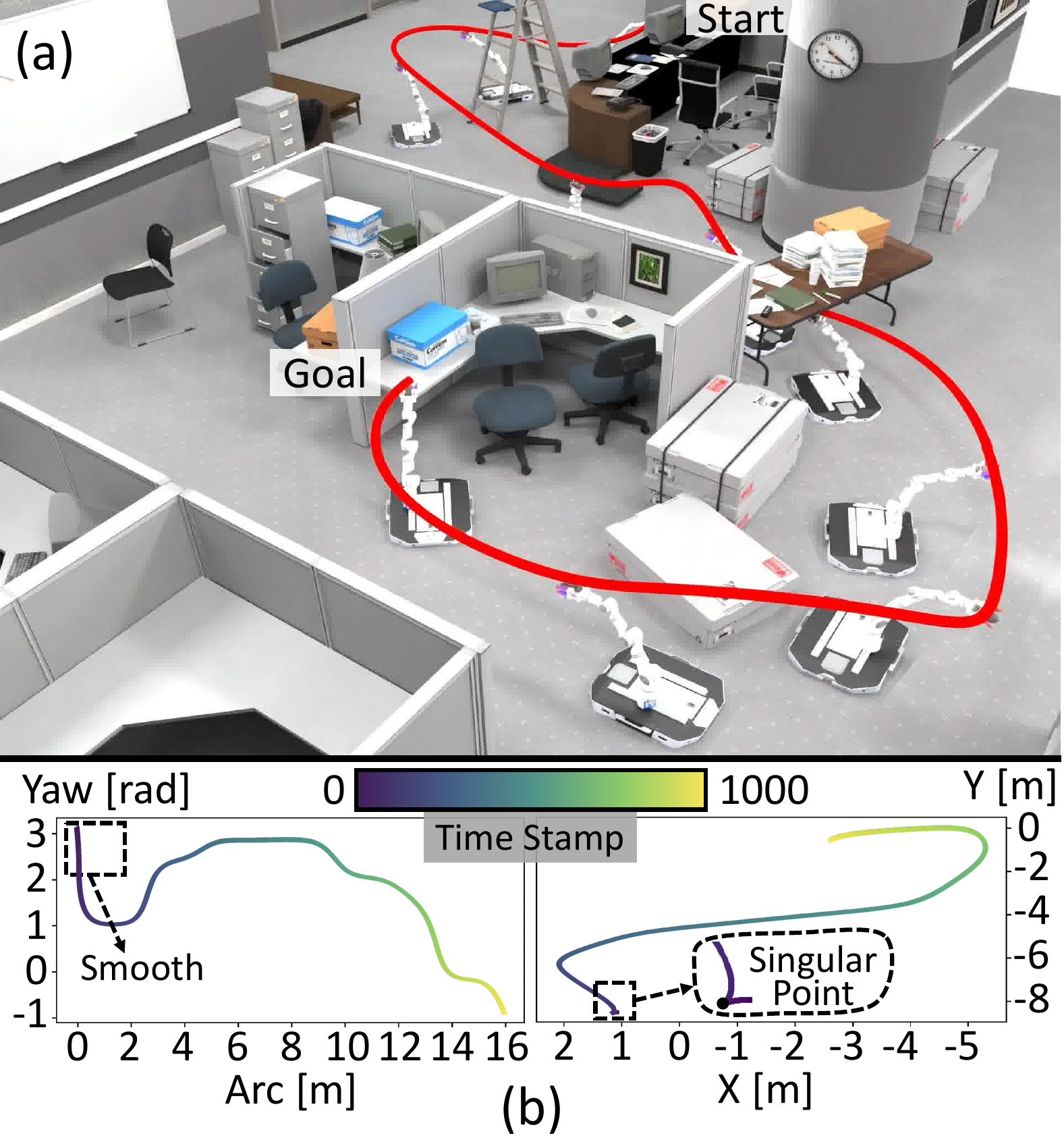}
	\caption{A differential drive mobile manipulator delivers a mouse from the reception to a workstation in an office using the proposed planner. The red curve in 
    Figure (a) represents the trajectory of the gripper. Figure (b) shows the motion curves of the base in arc length-yaw and Cartesian space.}\label{fig:head}
    \vspace{-0.3cm}
\end{figure}


Intuitively, the set of feasible paths for the DDB component within a DDMoMa is a subset of the feasible paths for a standalone DDB.
Leveraging this, we propose a hierarchical initial value acquisition algorithm, introducing topological paths search for the DDB. Conditioned on the DDB paths found, parallelized state sampling for the manipulator is then conducted. In a decoupled manner, the computational burden associated with high dimensionality is largely mitigated, improving the efficiency and success rate for planning in complex scenarios.



For handling the high dimension of optimization variables due to the high-dimensional state space, we propose a novel trajectory representation for DDMoMa inspired by polynomial trajectory parameterization. This parameterization has recently proven highly effective in robot motion planning, owing to its intrinsic smoothness and its substantially reduced dimensionality compared with finite-element discretizations\cite{minco, zmk, hzc_sr}. To address nonholonomic constraints of the DDB, we introduce arc length-yaw parameterization, where arc length at time $t=\tau$ is defined as the signed distance from time $t=0$ to $t=\tau$ along the trajectory, as shown in Fig.~\ref{fig:head}(b). This representation demonstrates higher efficiency than the state-of-the-art (SoTA) method\cite{zzd_icra} that uses differential flatness\cite{zmkicra} in Cartesian space.


Integrating these components, we present TopAY, an efficient optimization-based trajectory planning framework for DDMoMa that integrates hierarchical initial value acquisition algorithm and proposed trajectory representation. We also implement parallel trajectory optimization to further improve efficiency. Comprehensive simulation and real-world experiments demonstrate the effectiveness and efficiency of our approach. The contributions of the paper are:

\begin{itemize}
	\item We propose an efficient hierarchical path acquisition method for DDMoMa by introducing topological path search for the base and parallelized path sampling for the manipulator.
	\item We propose a novel polynomial-based trajectory representation for DDMoMa that handles the nonholonomic kinematics of the differential drive base with arc length-yaw parameterization.
	\item Integrating the above two modules, we present an efficient optimization-based trajectory planning framework for DDMoMa. Parallel trajectory optimization is deployed to achieve higher efficiency.
\end{itemize}

\section{Related Works}
\label{sec:Related Work}
\subsection{Sampling-based Path Planning}
Sampling-based methods\cite{sampling} approximate the configuration space by sampling discrete waypoints uniformly or via heuristics-guided strategies\cite{irrts}. They incrementally construct graphs where nodes represent sampled waypoints and edges denote collision-free paths. Feasible paths can then be derived by querying the graphs. This kind of method typically provides two key theoretical guarantees: probabilistic completeness and asymptotic optimality. However, the performance severely degrades in high-DoF systems, such as mobile manipulators\cite{zzd_icra}. The exponential growth of sampling space volume with dimensionality results in excessively long planning time in challenging scenarios, limiting the practicality.

One approach to improve efficiency involves employing alternative sampling strategies informed by prior knowledge of the workspace. Since only a subset of sampled waypoints contribute to the final path, it is natural to prioritize sampling in regions where these waypoints are more likely to be found with respect to the workspace and the planning task, i.e. using biased sampling distributions conditioned on the workspace\cite{learn_sample_begin,learn_dataset,learn_neural,mp_net}. An advantage of this approach is its ability to preserve both guarantees of completeness and optimality by combining uniform distribution with the biased distribution. The challenge lies in balancing the cost and frequency of sampling or iteration, as obtaining biased samples often requires more time. 

The alternative approach\cite{zzd_icra,rampage} mitigates the performance degradation induced by high DoF of the mobile manipulator by decomposing the manipulator and base motions, one conditioned on the other. Although this strategy sacrifices completeness, it greatly improves the efficiency of planning for DDMoMa and achieves real-time performance.

Recent work has also employed diffusion models to sample directly in the trajectory space\cite{m2diffuser}, thereby bypassing the iterative graph-construction process. Its primary limitations include slow training and inference, as well as a strong dependence on the design of the objective function.

\subsection{Optimization-based Trajectory Generation}
When higher-order robot states, such as acceleration and angular acceleration, need to be considered to obtain better solutions, sampling-based methods need to search in a larger space, whereupon the problem of combinatorial explosion is magnified, severely affecting the efficiency. Optimization-based methods tackle this limitation by formulating motion planning as an optimization problem to generate smooth, optimal, and dynamically feasible trajectories. In practice, sampling-based path planning is often utilized as the front-end to initialize the back-end trajectory optimization. 

The high DoF in DDMoMa renders optimization computationally expensive and poses a significant challenge to real-time planning. To ensure tractability, one strategy is receding horizon planning\cite{rampage,localmpc}. For example, RAMPAGE\cite{rampage} formulates motion generation as an integrated planning and control problem, planning over a receding horizon to substantially reduce computational cost. However, its inability to account for long-term consequences could lead to myopic behavior, such as getting trapped in dead ends.

To enable real-time global planning, REMANI\cite{zzd_icra} leverages the MINCO trajectory class\cite{minco} to reduce the number of optimization variables, greatly accelerating the trajectory planning for DDMoMa. However, the optimization process is affected by numerical issues arising from singularities in differential flatness they use for trajectory parameterization. Moreover, since a greedy strategy is adopted in path search, it fails to provide diverse initialization sets for trajectory optimization, harming its overall efficiency in complex scenarios.

Our method is also optimization-based. Unlike REMANI\cite{zzd_icra}, we represent DDB trajectories in the arc length-yaw space to avoid singularities from differential flatness. We further apply topological path search for DDMoMa path sampling, which diversifies initializations and improves success rates in complex scenarios. Moreover, inspired by cuRobo’s use of CUDA\cite{curobo} for efficient manipulator trajectory generation and inverse kinematics, we parallelize both the manipulator and trajectory optimization modules to boost efficiency.

\section{Planning Framework}
\label{sec:Planning Framework}
\begin{figure}[t]
    \centering
    \includegraphics[width=1.0\columnwidth]{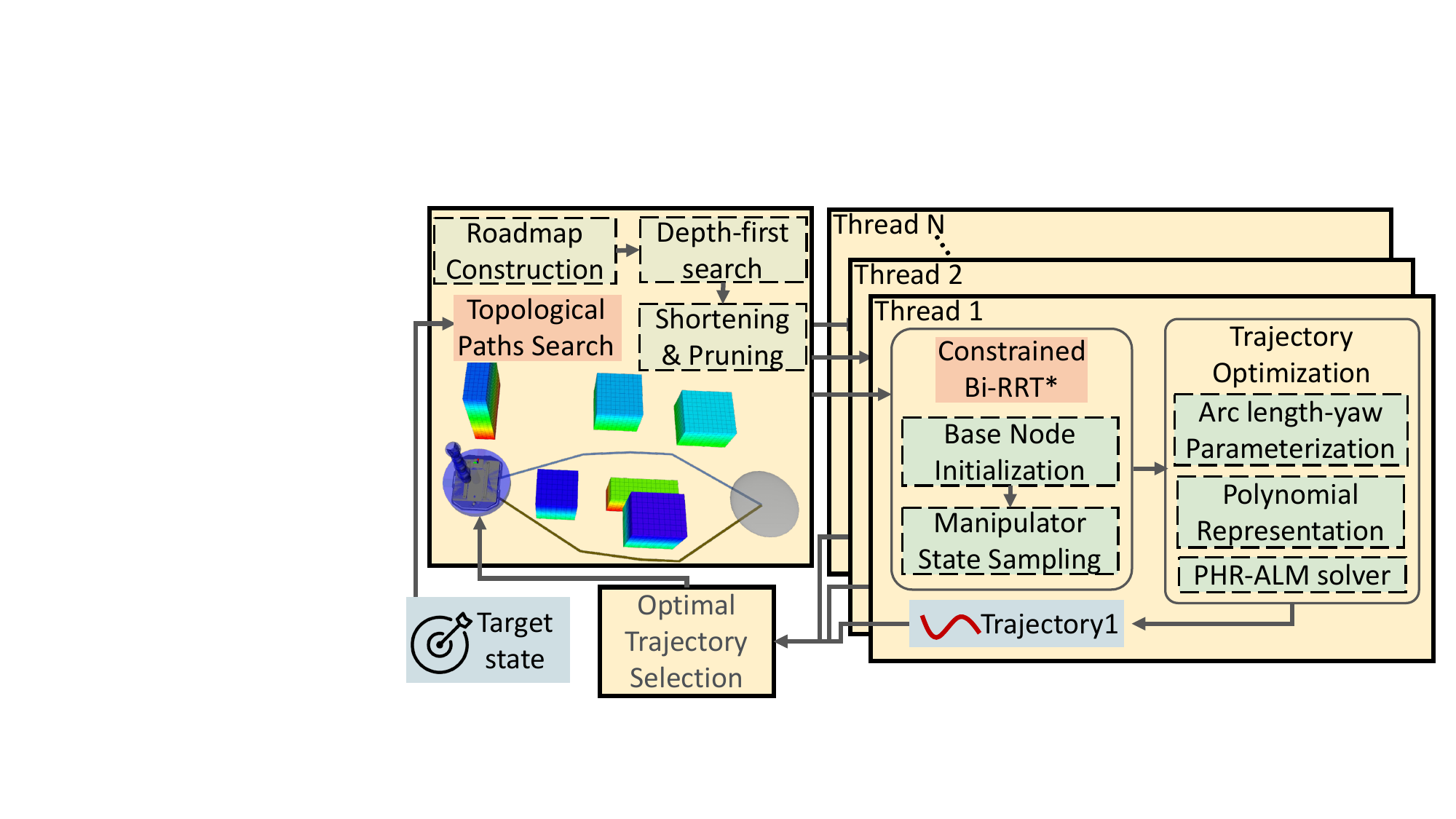}
	\caption{Planning framework.}\label{fig:framework}
\end{figure}

Fig.~\ref{fig:framework} illustrates our planning framework. When the robot receives the target $SE(3)$ pose of the end effector, it performs topological paths search to obtain different 2D paths for the DDB. The planner then initializes the sequence of the base states in $SE(2)$ for these paths and samples the manipulator states to obtain whole-body paths of the robot in parallel, which are subsequently handed over to the trajectory optimizer to obtain dynamically feasible trajectories.

\subsection{Hierarchical Path Acquisition}
Hierarchical path acquisition for DDMoMa mainly consists of topological paths search and constrained bi-directional informed RRT*, as shown in the orange-marked parts in Fig.~\ref{fig:framework}. The pipeline is shown in Algorithm~\ref{alg:searching}. The idea of topological paths has been used in many works\cite{topo_tits,zzd_topo,topo_tro,vprm} for motion planning. In this paper, we adopt the \textit{uniform visibility deformation} (UVD) class ~\cite{zzd_topo} to achieve efficient topological paths search in 2D environments (Lines \ref{line:topb}-\ref{line:tope}). Specifically, a visibility probabilistic roadmap (PRM)\cite{vprm} is first constructed. A depth-first search (DFS) is then applied to this graph to discover a diverse set of paths. These paths are shortened in parallel by checking the visibility between nodes. Further categorising them into different UVD classes, sorting and filtering them based on path length, we obtain the final topological paths $\mathcal{P}_b$.

Using the set interval $\Delta l$, each topological path in $\mathcal{P}_b$ will be discretized into a sequence of states in $SE(2)$ (i.e. $\boldsymbol{P}_{b}$), which is then used as constraints of path sampling for the manipulator (Lines \ref{line:mcrrtb}-\ref{line:mcrrte}). Here, we combine RRT-connect\cite{rrtconnect} and iRRT*\cite{irrts} for sampling, node expansion, rewiring, tree switching, connection attempts, and merging. In this work, $\boldsymbol{P}_{b}$ imposes the following constraints on the algorithm: 1) The base states of the nodes from sampling or in trees $\mathcal{T}_a,\mathcal{T}_b$ are restricted to discrete sequence points $\boldsymbol{P}_b$. 2) When searching for neighbouring nodes ($\mathtt{Nearest}$) and expanding nodes ($\mathtt{Steer}$), the base state is selected from the adjacent $SE(2)$ states in the corresponding sequence.

\begin{algorithm}
    \caption{Hierarchical Path Acquisition for differential drive mobile manipulator}
    \label{alg:searching}
    \KwIn{grid map $\mathcal{M}$, start and target state $\boldsymbol{s}_0,\boldsymbol{s}_f\in SE(2)\times\mathbb{R}^\text{N}$, constants $\Delta l,t_\text{smax}$.}
    \KwOut{paths of DDMoMa $\mathcal{P}$}
    $\mathcal{G}\leftarrow\mathtt{CreatePRM}(\boldsymbol{s}_0,\boldsymbol{s}_f,\mathcal{M})$\;\label{line:topb}
    $\mathcal{P}_b^\text{raw}\leftarrow\mathtt{DFS}(\boldsymbol{s}_0,\boldsymbol{s}_f,\mathcal{G})$\;
    $\mathcal{P}_b^\text{sc}\leftarrow\mathtt{ParallelShortCut}(\mathcal{P}_b^\text{raw},\mathcal{M})$\;
    $\mathcal{P}_b\leftarrow\mathtt{PruneUVD}(\mathcal{P}_b^\text{sc},\mathcal{M});\mathcal{P}\leftarrow\varnothing$\;\label{line:tope}
    \textcolor[rgb]{0.086,0.38,0.67}{\#pragma parallel }\textbf{for}\\
    \For{\textbf{each} $\boldsymbol{P}_r\in\mathcal{P}_b$}
    {
        $\boldsymbol{P}_b\leftarrow\mathtt{DiscreteSE2}(\boldsymbol{P}_r,\Delta l)$\;\label{line:mcrrtb}
        $\mathcal{T}_a.\mathtt{init}(\boldsymbol{s}_0);\mathcal{T}_b.\mathtt{init}(\boldsymbol{s}_f)$\;
        $\boldsymbol{n}_a,\boldsymbol{n}_b\leftarrow\mathtt{NULL};c_{max}\leftarrow+\infty$\;
        \While{$\neg\ \mathtt{TimeOut}(t_\text{smax})$}
        {
            $\mathtt{TrySwap}(\mathcal{T}_a,\mathcal{T}_b);\boldsymbol{n}_r\leftarrow\mathtt{Sample}(\boldsymbol{P}_b,c_{max})$\;
            $\boldsymbol{n}_n\leftarrow\mathtt{Nearest}(\mathcal{T}_a,\boldsymbol{n}_r)$\;
            $\boldsymbol{n}_s\leftarrow\mathtt{Steer}(\boldsymbol{n}_n,\boldsymbol{n}_r)$\;
            \lIf{$\mathtt{IsInvalid}(\boldsymbol{n}_s)$}{ \textbf{continue} }
            \uIf{$\boldsymbol{n}_s\in\mathcal{T}_b$}
            {
                \textcolor[rgb]{0.13,0.65,0.458}{// try updating $c_{max}$, i.e. $\mathtt{TryUpdateCost}$}\\
                $c=\mathtt{Cost}(\boldsymbol{n}_n)+\mathtt{Cost}(\boldsymbol{n}_s)+\mathtt{Heu}(\boldsymbol{n}_n,\boldsymbol{n}_s)$\;
                \uIf{$c_{max}>c$}
                {
                    $c_{max}\leftarrow c;\boldsymbol{n}_a\leftarrow\boldsymbol{n}_n;\boldsymbol{n}_b\leftarrow\boldsymbol{n}_s$\;
                    \textbf{continue}\;
                }
            }
            \uIf{$\mathtt{JustExpand}(\boldsymbol{n}_s)\lor(\mathtt{Cost}(\boldsymbol{n}_s)>\mathtt{Cost}(\boldsymbol{n}_n)+\mathtt{Heu}(\boldsymbol{n}_n,\boldsymbol{n}_s))$}
            {
                $\mathtt{Link}(\boldsymbol{n}_n,\boldsymbol{n}_s);\mathtt{Rewire}(\mathcal{T}_a,\boldsymbol{n}_s)$\;
                $\boldsymbol{n}_{on}\leftarrow\mathtt{Nearest}(\mathcal{T}_b,\boldsymbol{n}_s)$\;
                $\boldsymbol{n}_{os}\leftarrow\mathtt{Steer}(\boldsymbol{n}_{on},\boldsymbol{n}_s)$\;
                \lIf{$\mathtt{IsInvalid}(\boldsymbol{n}_{os})$}{ \textbf{continue} }
                \uIf{$\boldsymbol{n}_{os}\in\mathcal{T}_a$}
                {
                    $\mathtt{TryUpdateCost}(\boldsymbol{n}_{on},\boldsymbol{n}_{os},c_{max})$\;
                }
                \uIf{$\mathtt{JustExpand}(\boldsymbol{n}_{os})\lor(\mathtt{Cost}(\boldsymbol{n}_{os})>\mathtt{Cost}(\boldsymbol{n}_{on})+\mathtt{Heu}(\boldsymbol{n}_{on},\boldsymbol{n}_{os}))$}
                {
                    $\mathtt{Link}(\boldsymbol{n}_{on},\boldsymbol{n}_{os});\mathtt{Rewire}(\mathcal{T}_b,\boldsymbol{n}_{os})$\;
                    $\mathtt{TryConnect}(\boldsymbol{n}_{os},\mathcal{T}_a)$\;
                }
            }
        }
        \uIf{$\mathtt{IsValid}(\boldsymbol{n}_a)\land\mathtt{IsValid}(\boldsymbol{n}_b)$}
        {
            $\mathtt{MergeTree}(\mathcal{T}_a,\mathcal{T}_b,\boldsymbol{n}_a,\boldsymbol{n}_b)$\;
            $\mathcal{P}.\mathtt{Push}(\mathtt{GetPath}(\mathcal{T}_a,\mathcal{T}_b))$\;\label{line:mcrrte}
        }
    }
    \textbf{return} $\mathcal{P}$.
\end{algorithm}

\subsection{Trajectory Representation}
Let the $SE(2)$ state trajectory of the DDB be $[x(t),y(t),\theta(t)]^\text{T}$. In work~\cite{zzd_icra}, Wu et al. use differential flatness to handle the nonholonomic kinematics of DDB, where $\dot x(t)=v(t)\cos\theta(t)$, $\dot y(t)=v(t)\sin\theta(t)$, $\dot\theta(t)=\omega(t)$. They adopt $x(t)$ and $y(t)$ in Cartesian space to represent the DDB trajectory, explicitly deriving the velocity $v(t)=\eta\sqrt{\dot x^2+\dot y^2}$, yaw $\theta(t)=\text{atan2}(\eta\dot y,\eta\dot x)$, and angular velocity $\omega(t)=(\dot x\ddot y-\dot y\ddot x)/(\dot x^2+\dot y^2)$ of the DDB, achieving efficient trajectory planning for DDMoMa, where $\eta=1\lor-1$ denotes the DDB is moving forward or backward.

However, when the DDB is stationary or is spinning in place, i.e., when $v(t) = 0$, the function $\text{atan2}$ in the formula for yaw will be undefined, and the denominator in the formula for angular velocity will be zero, which is known as the singularity of differential flatness\cite{hzc_sr}. 
This singularity causes numerical instabilities that prevent the satisfication of the DDB's angular velocity and angular acceleration constraints. To alleviate this, existing methods impose workarounds like a minimum velocity constraints\cite{zmkicra} or dense constraint points near singular points \cite{zzd_icra}, which inadvertently reduce planning efficiency.
Moreover, such parameterization method is unable to represent spinning-place-maneuvers, where $\dot x(t) = \dot y(t) = 0$, resulting in undefined yaw and angular velocity.

To address this problem, in this work, we introduce a novel trajectory representation based on arc length $s(t)$ and yaw angle $\theta(t)$, as shown in Fig.~\ref{fig:head}(b). In which, the position is given by integration of the kinematics equations:
\begin{align}
x(t)&=\int_0^t\dot s(\tau)\cos\theta(\tau)d\tau+x_0,\label{intx}\\
y(t)&=\int_0^t\dot s(\tau)\sin\theta(\tau)d\tau+y_0,\label{inty}
\end{align}
where $[x_0,y_0]^\text{T}$ is the initial position of the DDB. Following the optimality condition for multi-stage control effort minimization problem\cite{minco}, we use quintic piecewise polynomials with four times continuously differentiable at the segmented points to represent the trajectory of DDMoMa for minimizing jerk\cite{curobo}. Each piece of trajectory is denoted as:
\begin{align}
s_{j}(t)&=\boldsymbol{c}_{s_j}^\text{T}\gamma(t)\quad &t\in[0,T_{j}],\\
\theta_{j}(t)&=\boldsymbol{c}_{\theta_j}^\text{T}\gamma(t)\quad &t\in[0,T_{j}],\\
q_{kj}(t)&=\boldsymbol{c}_{q_{kj}}^\text{T}\gamma(t)\quad &t\in[0,T_j],
\end{align}
where $q_{kj}(t)$ is the trajectory of the manipulator. $k\in\mathbb{Z}\cap[1,\text{N}]$ is the index of the joints, $j\in\mathbb{Z}\cap[1,\text{M}]$ is the index of piecewise polynomial. $T_j$ is the duration of a piece of the trajectory. $c_*\in\mathbb{R}^6,*=\{s_j,\theta_j,q_{kj}\}$ is the coefficient of polynomial. $\gamma(t)=[1,t,t^2,...,t^5]^\text{T}$ is the natural base. In this work, we employ Simpson’s rule to calculate the integrals (\ref{intx}) and (\ref{inty}) numerically.

\subsection{Trajectory Optimization}
In this paper, we formulate the trajectory optimization problem for DDMoMa as:
\begin{align}
\min_{\boldsymbol{c},\boldsymbol e,\boldsymbol{T}}&f(\boldsymbol{c},\boldsymbol e,\boldsymbol{T})=\int_0^{\sum_{j=1}^\text{M}T_j}\boldsymbol{j}(t)^\text{T}\boldsymbol W\boldsymbol{j}(t)dt+\rho\lVert\boldsymbol{T}\rVert_1\label{problem:opt}\\
&s.t.\ \ \textbf{FK}(\boldsymbol{s}_f)=\boldsymbol{p}_{e},\label{con:fk}\\
&\quad\quad \boldsymbol{M}(\boldsymbol{T})\boldsymbol{c}=\boldsymbol{b}(\boldsymbol{P},\boldsymbol{e}),\quad \boldsymbol{T}>\boldsymbol{0},\label{con:minco}\\
&\quad \quad \lvert v_\text{max}\omega(t)\pm\omega_\text{max}v(t)\rvert\leq v_\text{max}\omega_\text{max},\label{con:dyn_begin}\\
&\quad \quad a^2(t)\leq a_\text{max}^2,\quad \beta^2(t)\leq\beta_\text{max}^2,\\
&\quad \quad (\boldsymbol{q}\circ\boldsymbol{q})(t)\leq \boldsymbol{q}_\text{max2},\quad(\dot{\boldsymbol{q}}\circ\dot{\boldsymbol{q}})(t)\leq \dot{\boldsymbol{q}}_{\text{max2}},\\
&\quad \quad (\ddot{\boldsymbol{q}}\circ\ddot{\boldsymbol{q}})(t)\leq \ddot{\boldsymbol{q}}_{\text{max2}},\label{con:dyn_end}\\
&\quad \quad \textbf{SDF}(\textbf{ColliPts}(\boldsymbol{c},\boldsymbol{T}))\geq\boldsymbol r_{thr},\label{con:safee}\\
&\quad \quad \textbf{SelfColli}(\boldsymbol{c},\boldsymbol{T})\ge\boldsymbol{0},\label{con:safes}
\end{align}
where 
$\boldsymbol{c}=[\boldsymbol{c}_{1}^\text T,\boldsymbol{c}_{2}^\text T,...,\boldsymbol{c}_{\text{M}}^\text T]^\text T\in\mathbb{R}^{6\text{M}\times(\text{N+2})}$, $\boldsymbol{c}_{k}=[\boldsymbol{c}_{k1},\boldsymbol{c}_{k2},...,\boldsymbol{c}_{k(\text{N+2})}]\in\mathbb{R}^{6\times(\text{N+2})},k=1,2,...,\text{M}$ are coefficient matrices. For other optimization variables, $\boldsymbol{s}_f=[x_f,y_f,\theta_f,\boldsymbol{q}_f]^\text T$, $\boldsymbol{T}=[T_1,T_2,..,T_\text{M}]^\text{T}$ denote the end state of the robot and time durations of the trajectory, respectively, where $\boldsymbol{q}_f=[q_{1f},q_{2f},...,q_{\text{N}f}]$. $x_f,y_f$ are calculated using $\boldsymbol{e}=[s_f,\theta_f,\boldsymbol{q}]^\text{T}$ and $\boldsymbol{c},\boldsymbol{T}$. All equations and inequalities are taken element-wise.

In the objective function, $\boldsymbol{j}(t)=[s^{(3)}(t),\theta^{(3)}(t),\boldsymbol{q}^{(3)}(t)]^\text T$ denotes the jerk of the trajectory. $\boldsymbol W\in\mathbb{R}^{(2+\text{N})\times(2+\text{N})}$ is a diagonal positive matrix representing the weights. $\rho>0$ is a constant to give the trajectory some aggressiveness.

Eq.(\ref{con:fk}) denotes the end-effector pose constraints for the robot end state. $\textbf{FK}:SE(2)\times\mathbb{R}^{\text{N}}\mapsto SE(3)$ is the forward kinematics function. Conditions (\ref{con:minco}) denotes the combinations of the continuity constraints mentioned in last subsection and boundary conditions of the trajectory:
\begin{align}
[s(0),\dot s(0),\ddot s(0)]&=[0,v_0,a_0],\\
[\theta(0),\dot \theta(0),\ddot\theta(0)]&=[\theta_0,\omega_0,\beta_0],\\
[\boldsymbol{q}(0),\dot{\boldsymbol{q}}(0), \ddot{\boldsymbol{q}}(0)]&=[\boldsymbol{q}_0,\dot{\boldsymbol{q}}_0,\ddot{\boldsymbol{q}}_0],\\
[s(T_f),\dot s(T_f),\ddot s(T_f)]&=[s_f,0,0],\\
[\theta(T_f),\dot \theta(T_f),\ddot\theta(T_f)]&=[\theta_f,0,0],\\
[\boldsymbol{q}(T_f),\dot{\boldsymbol{q}}(T_f), \ddot{\boldsymbol{q}}(T_f)]&=[\boldsymbol{q}_f,\boldsymbol{0},\boldsymbol{0}].
\end{align}
$\boldsymbol{P}\in\mathbb{R}^{(\text{N}+2)\times(\text M-1)}$ is the segment points matrix. $T_f=\lVert\boldsymbol{T}\rVert_1$ is the total duration of the trajectory.

Conditions (\ref{con:dyn_begin})$\sim$(\ref{con:dyn_end}) are dynamic feasibility constraints, including coupled linear velocity $v(t)=\dot s(t)$ and angular velocity $\omega(t)=\dot\theta(t)$ constraints\cite{zmk}, limitation of linear acceleration $a(t)=\dot v(t)$, angular acceleration $\beta(t)=\dot\omega(t)$, joint angles, joint angular velocities and joint angular accelerations, where $v_\text{max},\omega_\text{max},a_\text{max},\beta_\text{max},\boldsymbol{q}_{\text{max2}}=\boldsymbol{q}_\text{max}\circ\boldsymbol{q}_\text{max},\dot{\boldsymbol{q}}_{\text{max2}}=\dot{\boldsymbol{q}}_\text{max}\circ\dot{\boldsymbol{q}}_\text{max},\ddot{\boldsymbol{q}}_{\text{max2}}=\ddot{\boldsymbol{q}}_\text{max}\circ\ddot{\boldsymbol{q}}_\text{max}$ are constants or constant vectors. The operator $\circ:\mathbb{R}^\text{N}\times\mathbb{R}^\text{N}\mapsto\mathbb{R}^\text{N}$ denotes the Hadamard product.

Condition (\ref{con:safee}) denotes constraints related to obstacle avoidance. In this paper, we construct a Signed Distance Field (SDF) to represent the environment. In SDF, the value at each state of the space is the distance to the edge of the nearest obstacle, which is negative inside the obstacles. As shown by the robot collision model in Fig.~\ref{fig:implement}, we approximate the collision geometry using a set of cylinders and spheres. The function $\textbf{ColliPts}(\cdot)$ maps the state points to sets of collision detection points, i.e., the centers of the cylinder and the spheres that constitute the collision model shown in Fig.\ref{fig:implement}. The function $\textbf{SDF}(\cdot)$ maps the set of points to the corresponding SDF values, where the SDF for the DDB is constructed with a 2D grid map. $\boldsymbol{r}_{thr}$ is a constant vector, which includes the radii of the cylinder and spheres. Condition (\ref{con:safes}) imposes self-collision avoidance constraints on the robot's configuration. Since our collision model consists of cylinder and spheres, the function $\textbf{SelfColli}(\cdot)$ calculates the distance vector representing the distance between each pair of them, corresponding to each state point of the trajectory.

\subsection{Problem Solving}

We address Conditions (\ref{con:minco}) using technique from existing work \cite{minco}, converting the optimization variables $\{\boldsymbol{c},\boldsymbol{e},\boldsymbol{T}\}$ to $\{\boldsymbol{P},\boldsymbol{e},\boldsymbol{\tau}\}$, where for each element $T_j$ in $\boldsymbol{T}$ and $\tau_j$ in $\boldsymbol{\tau}$, we have
\begin{align}
\tau_j=\text{L}_{c2}(T_j)=\left\{\begin{matrix}
1-\sqrt{2T_j^{-1}-1} & 0< T_j\le1\\
\sqrt{2T_j-1}-1 & T_j>1
\end{matrix}\right.\quad.
\end{align}

To guarantee that constraint $(\boldsymbol{q}\circ\boldsymbol{q})(t)\leq \boldsymbol{q}_\text{max2}$ is satisfied when $t = T_j,j\in\mathbb{Z}\cap[1,\text{M}]$ and at $\boldsymbol{q}_f$, we use the following inverse sigmoid-like function to convert $\boldsymbol{q}_f$ and each column of the submatrix $\boldsymbol{Q}\in\mathbb{R}^{\text{N}\times(\text{M}-1)}$ of $\boldsymbol{P}$ with respect to the joint angles of the manipulator to $\boldsymbol{\mathfrak{q}}_f$ and $\boldsymbol{\mathcal{Q}}_f$, respectively: 
\begin{align}
\mathfrak{q}(q)&=\text{L}_{c2}\left(\frac{q_\text{max}+q}{q_\text{max}-q}\right).
\end{align}

To deal with continuous polynomial trajectories, we discretize each piece of the duration $T_{j}$ as $\text K$ time stamps $\tilde{t}_{l}=(l/\text K)\cdot T_{j},l=0,1,...,K-1$, imposing the constraints on these time stamps. To efficiently solve this optimization problem, we convert all inequality constraints into penalty terms and add them to the objective function as discrete integrals. Specifically, let the objective function be $g(\text{traj})$, and one constraint function be $\mathcal{C}(\text{traj})$, then the new objective function is 
\begin{align}
g(\text{traj})+\rho_{c}\sum_{j=1}^{\text{M}}\sum_{l=0}^{K-1}\frac{T_j}{\text{K}}\mathcal{L}(\max\{0,\mathcal{C}[\text{traj}_j(\frac{lT_j}{\text{K}})]\}),
\end{align}
where $\mathcal{L}$ denotes a function that smooths the constraint function $\mathcal{C}$ into a $C^2$ function, $\text{traj}_j$ is the $j$-th polynomial trajectory. $\rho_c$ denotes the penalty weight. This technique has been demonstrated to be practically effective in numerous studies\cite{zzd_icra,minco,zmk,zmkicra}.

We employ Powell-Hestenes-Rockafellar Augmented Lagrangian method\cite{alm} (PHR-ALM) to handle the equation constraint (\ref{con:fk}). It combines Lagrange multipliers with a quadratic penalty to enforce constraints while maintaining good convergence properties. In this work, the efficient quasi-Newton optimizer L-BFGS\cite{lbfgs} is chosen as the inner loop iteration algorithm for PHR-ALM.


\section{Results}
\label{sec:Results}
\subsection{Implementation Details}
\begin{figure}[t]
    \centering
    \includegraphics[width=1.0\columnwidth]{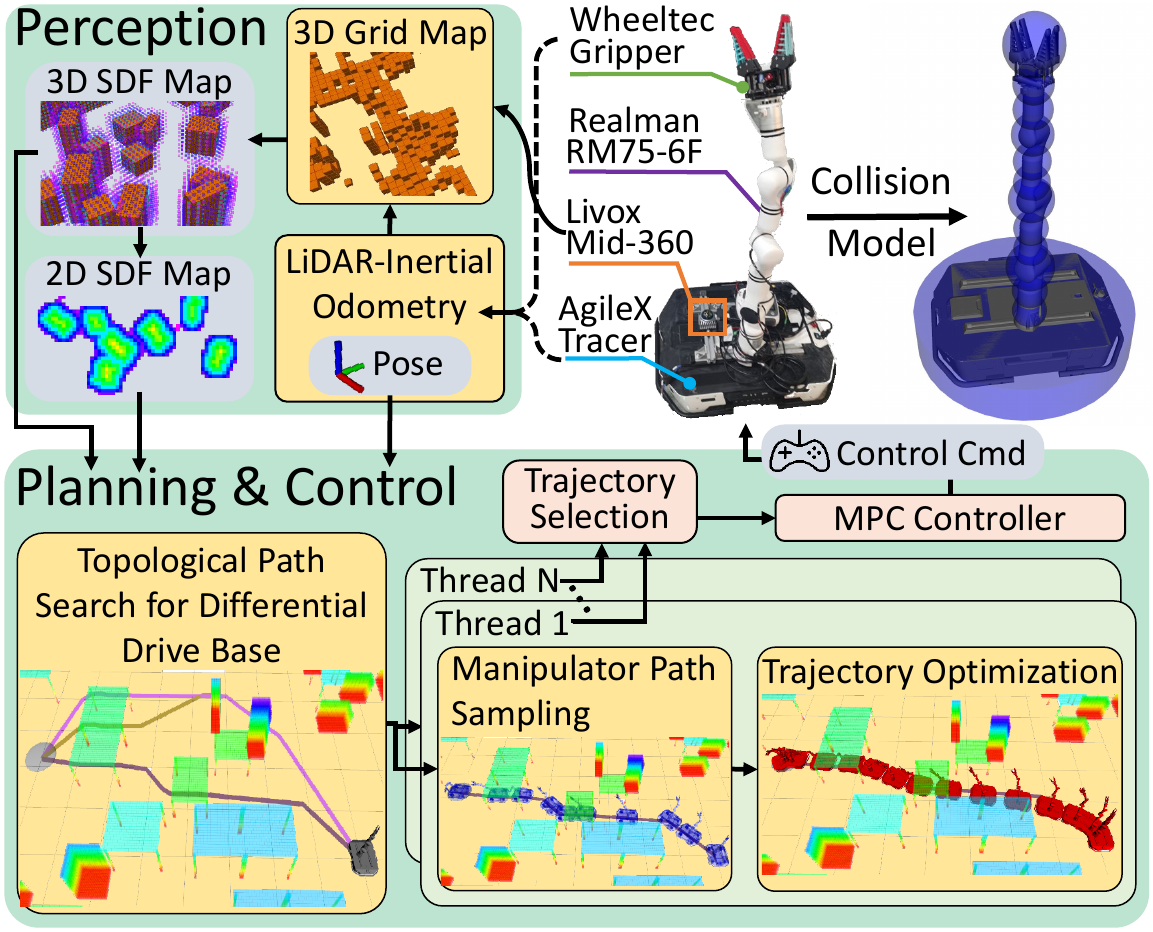}
	\caption{Hardware settings and software system of the DDMoMa used in real-world experiments.}\label{fig:implement}
\end{figure}

To validate the performance of our pipeline in real-world applications, we deploy it on a DDMoMa, as shown in the top right of Fig.~\ref{fig:implement}. All computations are performed by an onboard computer Intel NUC 11 Phantom Canyon. The software system of the robot primarily consists of three modules, perception, planning, and control, as shown in the green box area in Fig.~\ref{fig:implement}. In the perception module, we utilize FAST-LIO2\cite{fast_lio} for localization, and ROG-Map\cite{rog_map} for maintenance and live-updating of the 3D grid map and SDFs.

In the planning and control module, we use Boost\footnote{\url{https://www.boost.org/}} to launch and manage multithreading. We employ an early termination strategy that waits an additional period after the first successful outcome result is returned by any thread. The shortest optimized trajectory from all successful threads is then selected. The selected one will be fed into a controller based on model predictive control to acquire control commands consisting of velocity, angular velocity of the DDB, and joints velocity of the manipulator. 
In order to enable adaptability to dynamic environment, a replanning procedure is invoked upon detection of obstacles colliding with currently tracking trajectory.

We also build two different simulation environments based on ROS\footnote{\url{https://www.ros.org/}} to carry out comparative experiments in different scenarios. All simulations are run on Ubuntu 20.04 with an Intel i7-12700 CPU.

\subsection{Real-world Experiments}
\label{subsec:Real-world Experiments}
\begin{figure*}[t]  
		\centering
		{\includegraphics[width=1.0\linewidth]{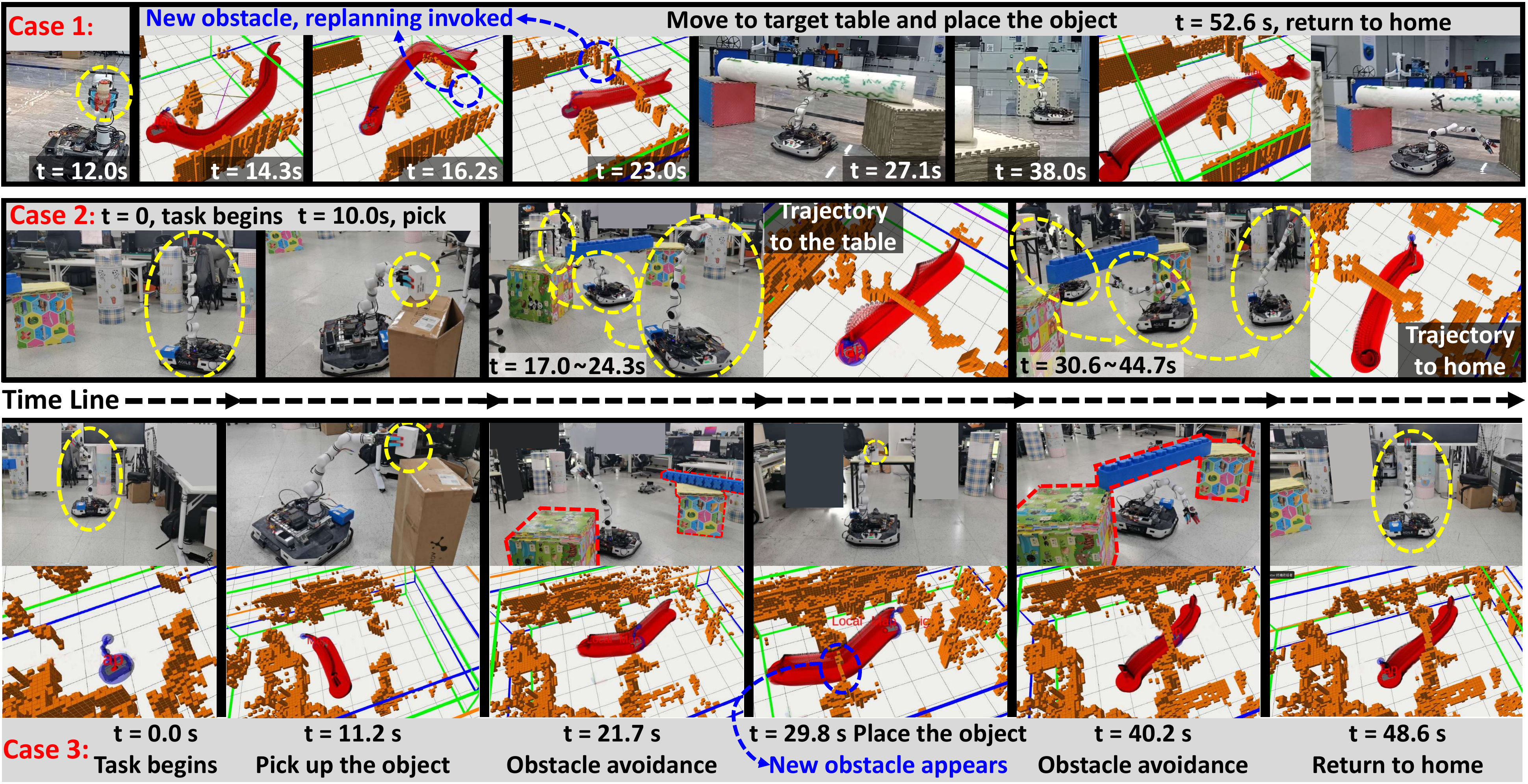}}
		\caption{Real-world experiments. The DDMoMa successfully performs pick-and-place tasks. Snapshots of filmed videos and RViz visualizations are presented in chronological order. In RViz visualizations, \textcolor[rgb]{0.87,0.42,0.0}{\textbf{orange}} boxes illustrate the 3D occupancy grid map being updated in real-time. The \textcolor[rgb]{0.23,0.21,1.0}{\textbf{blue}} silhouette represents the current state of the DDMoMa. The series of \textcolor[rgb]{0.8,0.04,0.004}{\textbf{red}} silhouettes represent the trajectory currently tracked by the DDMoMa.}
		\label{fig:real_exp}
\end{figure*}

For our real-world experiments, we tasked the DDMoMa with classical pick-and-place operation in three different indoor environments, as shown in Fig.~\ref{fig:real_exp}. 
For these experiments, the DDMoMa's task requires it to navigate among obstacles and involves reaching a designated area, grasping an object, delivering it to another area, and finally returning to its initial position.

In each experiment, the robot has no prior knowledge of the environment, apart from the initial position and drop-off location of the object, and thus relies on onboard sensor and real-time replanning to adapt to dynamic environments. As shown in Case 1, at $t = 16.2s$ and $t = 23.0s$, the robot performs two replannings upon discovery of obstacles in its original trajectory. Ultimately, the robot successfully finds a collision-free path and traverses under the beam-like obstacle.

In case 3, we modify the environment in the process of its operation. A beam-like obstacle is placed on top of two cubic obstacles, forming a bridge-like structure that is clearly visible at $t=40.2s$. The robot exhibited adaptability to dynamic environment by successfully finding a trajectory to avoid newly placed obstacle in real-time by bending its manipulator.

This series of experiments demonstrates the effectiveness and real-time performance of our algorithm. More demonstrations with static scenarios can be found in the attached multimedia. In these cases, we randomly set the target state of the robot to force it to continuously traverse the obstacle area, demonstrating the practicality of the method.

In real-world experiments, the limits of linear velocity, linear acceleration, angular velocity, and angular acceleration of the DDB are set to $v_\text{max}=1.0m/s$, $a_\text{max}=0.8m/s^2$, $\omega_\text{max}=0.9rad/s$, and $\beta_\text{max}=1.0rad/s^2$, respectively. The limits of joint angles, joint angular velocities, and joint angular accelerations of the manipulator are set to $\boldsymbol{q}_\text{max}=[3.1, 2.26, 3.1, 2.355, 3.1, 2.23, 6.28]^\text{T}$, $\dot{\boldsymbol{q}}_\text{max}=\mathtt{ConstantVec}(2.35)$, and $\ddot{\boldsymbol{q}}_\text{max}=\mathtt{ConstantVec}(6.28)$, where the units are $rad$, $rad/s$, $rad/s^2$, respectively. The function $\mathtt{ConstantVec}:\mathbb{R}\mapsto\mathbb{R}^7$ constructs a constant vector from a constant. For safety, we set $\boldsymbol{r}_{thr}=[40,6,6,8,4,4,7,3.5,3.5,6,3.5,3.5,8]$, where the unit is centimeter, and the order corresponds to the geometric solids from bottom to top in the collision model in Fig.~\ref{fig:implement}.

\subsection{Simulation Experiments}
\label{subsec:Simulation Experiments}
\begin{figure*}[t]
    \centering
    \includegraphics[width=1.0\linewidth]{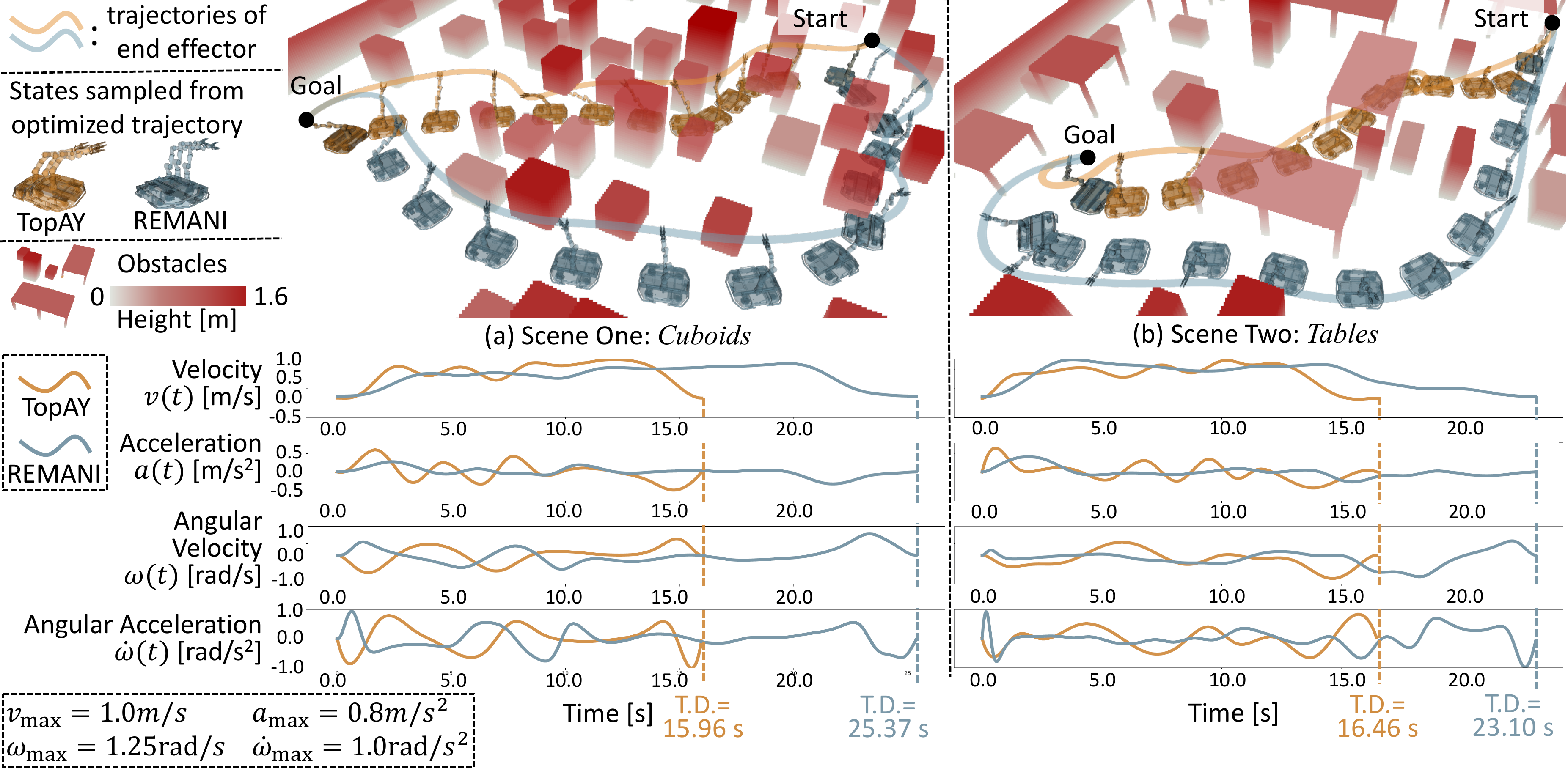}
	\caption{Motion trajectories of the DDMoMa (upper half) and time plots of the DDB's kinematic variables (lower half).}\label{fig:sim_exp}
\end{figure*}

We employ two different scenes in simulation for ablation study and benchmark comparisons, which are denoted by \textit{Cuboids} and \textit{Tables}, as shown in Fig.~\ref{fig:sim_exp} (a) and (b), respectively. Both are $20m \times 20m$ walled rooms filled with randomly placed obstacles of random size. For \textit{Cuboids}, there are 80 grounded cuboids and 80 floating cuboids. For \textit{Tables}, there are 80 grounded cuboids and 40 tables with rows and columns of legs. 
\textit{Tables} is considered a more challenging scene, as cluttered tables with nonconvex structure form many narrow passages and the dense distribution of table legs makes the DDMoMa more susceptible to collision.

\begin{table}[t]
\small
\centering
\renewcommand\arraystretch{1.2}
\caption{Ablation Studies. In each scene, \textbf{bold} or \underline{underline} indicates TopAY achieves the \textbf{best} or the \underline{second best} result.}
\label{tab:ablation}
\begin{tabular}{c|c|ccc}
\hline
Scene                    & Method     & S.R. (\%)     & T.P. (ms)      & T.D. (s)   \\ \hline
\multirow{3}{*}{Cuboids} & TopAY   & {\ul 94.4}    & \textbf{218.0} & {\ul 13.2} \\
                         & JPSOnly    & 88.1          & 226.3          & 13.7       \\
                         & Sequential & 95.1          & 783.3          & 13.0       \\ \hline
\multirow{3}{*}{Tables}  & TopAY   & \textbf{92.5} & {\ul 312.0}    & {\ul 15.6} \\
                         & JPSOnly    & 88.4          & 295.0          & 15.6       \\
                         & Sequential & 92.2          & 1363           & 15.3       \\ \hline
\end{tabular}
\end{table}

\textit{1) Ablation Studies:} Ablation studies are conducted to demonstrate the efficacy of topological paths search and parallel processing in proposed pipeline. With the starting position fixed at the center of the room, we randomly generate 1,000 goal states and conduct planning with three distinct methods: (1) TopAY, (2) TopAY, where the topological path search is replaced by Jump Point Search (JPSOnly), (3) TopAY without performing parallel manipulator path sampling and trajectory optimization. (Sequential)

We record the success rate (S.R.), average time consuming of planning (T.P.) and average trajectory duration (T.D.), as summarized in Table~\ref{tab:ablation}. Planning attempts will be considered failed if either the planner cannot find feasible path for DDB and manipulator within limited time or none of the optimization processes successfully return with dynamically feasible and collision-free trajectories. Only cases where all planners succeed contibute to T.P. and T.D. .

\begin{table*}[t]
\small
\centering
\renewcommand\arraystretch{1.3}
\caption{\label{tab:comparision} Algorithm Comparisons}
\begin{tabular}{c|cccccccccc}
\hline
\multirow{2}{*}{Scene} &
  Scale &
  \multicolumn{3}{c}{Small (3m $\sim$ 8m)} &
  \multicolumn{3}{c}{Medium (8m $\sim$15m)} &
  \multicolumn{3}{c}{Large (15m $\sim$ 30m)} \\ \cline{2-11} 
 &
  Method &
  S.R. (\%) &
  T.P. (ms) &
  T.D. (s) &
  S.R. (\%) &
  T.P. (ms) &
  T.D. (s) &
  S.R. (\%) &
  T.P. (ms) &
  T.D. (s) \\ \hline
\multirow{2}{*}{Cuboids} &
  \multicolumn{1}{c|}{TopAY} &
  \textbf{98.0} &
  \textbf{225.4} &
  \textbf{11.4} &
  \textbf{98.2} &
  \textbf{382.6} &
  \textbf{18.1} &
  \textbf{97.0} &
  \textbf{581.1} &
  \textbf{25.5} \\
 &
  \multicolumn{1}{c|}{REMANI\cite{zzd_icra}} &
  88.1 &
  467.6 &
  17.1 &
  81.4 &
  845.5 &
  24.1 &
  71.1 &
  1491 &
  31.8 \\ \hline
\multirow{2}{*}{Tables} &
  \multicolumn{1}{c|}{TopAY} &
      \textbf{87.5} &
  \textbf{225.6} &
  \textbf{12.4} &
  \textbf{74.1} &
  \textbf{417.6} &
  \textbf{20.0} &
  \textbf{61.8} &
  \textbf{794.7} &
  \textbf{28.6} \\
 &
  \multicolumn{1}{c|}{REMANI\cite{zzd_icra}} &
  55.1 &
  800.5 &
  18.0 &
  25.1 &
  1346 &
  26.5 &
  11.4 &
  4442 &
  37.5 \\ \hline
\end{tabular}
\end{table*}

TopAY achieves higher success rates compared to JPSOnly, owing to the enhanced probability of finding feasible subspaces for the manipulator through topological paths search. Similarly, this strategy makes it possible for optimizer to obtain better initial values, thereby reducing the trajectory durations. 
Although introducing multiple initial values increases computational load, the efficiency gap remains within 6\% thanks to the  parallelization. TopAY achieves even higher efficiency in \textit{Cuboids} due to the adoption of the early termination strategy in parallel trajectory optimization.

Compared to Sequential, 
the proposed method significantly reduced T.P. through extensive parallelization, achieving approximately a threefold improvement. Meanwhile, the early termination strategy does not result in a noticeable decrease in success rate or optimality, with the gaps remaining within 1\% and 2\%, respectively.

\textit{2) Comparisons with} REMANI\cite{zzd_icra}: We compare ToPAY with the SoTA trajectory optimization method for DDMoMa, REMANI\cite{zzd_icra}. Using the same scenarios, evaluation metrics and failure criteria as the ablation studies, 1,000 planning tasks are randomly generated for each scene and different distance range of the base. In each case, the parameters of all algorithms are meticulously tuned for the best overall performance. Results are summarized in Table~\ref{tab:comparision}.


In \textit{Cuboids}, TopAY maintains exceptional S.R. (>95\%) in all distance ranges, while REMANI\cite{zzd_icra} exhibits noticeable deterioration over increasing distance, which is more pronounced in \textit{Tables}. This stems from the greedy initial value acquisition of REMANI\cite{zzd_icra}, which samples manipulator paths based on only one kinematically feasible path for the base. When this strategy fails, it falls back to full state space sampling. As the problem scale increases, the failure probability of the greedy strategy rises, and full state sampling also becomes more difficult, leading to a significant increase in failure rate within finite time.

In terms of planning efficiency, 
TopAY consumes less than half the time of REMANI\cite{zzd_icra}, and can even achieve a five-fold speedup in \textit{Tables}. We attribute this to the use of topological paths and parallel processing. In obstacle-dense environments, planners are highly susceptible to failures, potentially requiring multiple optimization attempts. REMANI\cite{zzd_icra} adopts a simple retry-until-success strategy with no mechanism to ensure diversity of initialization. In contrast, TopAY significantly accelerates this process by leveraging parallel optimization with diverse initial values derived from topological paths.

Table~\ref{tab:comparision} also indicates TopAY's advantage over REMANI\cite{zzd_icra} in terms of trajectory duration. We attribute this enhancement to the usage of topological paths and arc length-yaw parameterization. The former facilitates the exploration of a diverse set of initial values, potentially yielding better solutions than methods using single initial value. At a theoretical level, the later addresses the difficulty of enforcing the DDB's angular velocity and angular acceleration constraints induced by singularity of differential flatness. Fig.~\ref{fig:sim_exp} provides a more intuitive comparison of TopAY and REMANI\cite{zzd_icra} through the motion trajectories of the robot and the time plots of the DDB’s kinematic variables.

\section{Conclusion \& Limitations}
In this paper, we present TopAY, an optimization-based trajectory planner for DDMoMa. The framework combines a hierarchical initial value acquisition method with a polynomial trajectory representation based on arc length–yaw parameterization to address high-dimensional state spaces and nonholonomic constraints. Simulation experiments demonstrate that TopAY achieves higher planning efficiency and success rate in complex scenarios compared to state-of-the-art method.

Despite these promising results, several limitations remain to be addressed for broader application. Our pipeline acquires initial value in a decoupled manner, significantly accelerating the procecss. However, this approach sacrifices completeness guarantee. Furthermore, in challenging cases, our method can remain computationally expensive and may fail to find stable and efficient solution within a reasonable time limit.


Neural networks, as a powerful implicit representation, have been shown to be a promising technique in motion planning\cite{m2diffuser,hzc_sr}. In the future, we will try to incorporate neural networks to address above issues to improve robustness and applicability. Besides, to further exploit the advantages of the efficiency of our method, adapting it to multi-arm robot systems and different floating bases, such as bimanual mobile manipulators, legged manipulators, and humanoid robots, will also be taken into consideration.

\bibliography{references}

\end{document}